# Reliable Force Aggregation Using a Refined Evidence Specification from Dempster-Shafer Clustering

Johan Schubert
Department of Data and Information Fusion
Division of Command and Control Warfare Technology
Swedish Defence Research Agency
SE–172 90 Stockholm, Sweden
schubert@foi.se
http://www.foi.se/fusion/

**Abstract -** *In this paper we develop methods for selection of templates and use these templates to recluster an already performed Dempster-Shafer clustering taking into account intelligence to template fit during the reclustering phase. By this process the risk of erroneous force aggregation based on some misplace pieces of evidence from the first clustering process is greatly reduced. Finally, a more reliable force aggregation is performed using the result of the second clustering. These steps are taken in order to maintain most of the excellent computational performance of Dempster-Shafer clustering, while at the same time improve on the clustering result by including some higher relations among intelligence reports described by the templates. The new improved algorithm has a computational complexity of* $O(n^3 \log^2 n)$ *compared to* $O(n^2 \log^2 n)$ *of standard Dempster-Shafer clustering using Potts spin mean field theory.*

**Keywords:** Force aggregation, Dempster-Shafer theory, Dempster-Shafer clustering, specification, template

## 1 Introduction

When evidence about different and multiple events are mixed up we want to arrange them in groups where all evidence in one group refers to the same event. This can be done by Dempster-Shafer clustering [1], [2], [3], [4], [5] using the conflict of Dempster's rule in Dempster-Shafer theory [6], as a distance measure. All evidence in each group can then be fused separately. The idea of using the conflict of Dempster's rule for this purpose has received a growing interest. It was described in [7], [8] as one example of some practical uses of belief functions, and is the basis for a forthcoming article on handling conflicts between sensor reports in multi-target detection [9].

In this paper we use a specification method [10] closely associated with Dempster-Shafer clustering to obtain a detailed specification about which events each piece of evidence might refer to [11]. This is done in order to find a core of evidence within each cluster containing pieces of evidence with high chance of being correctly placed. This specification about reference is derived from incremental changes to a metaconflict function when one piece of evidence is extracted from the cluster where it was placed by the Dempster-Shafer clustering and put into another group. As Dempster-Shafer clustering uses only pairwise conflicts between pieces of evidence it is computationally efficient but can not take into account more complex relations in the evidence.

The core of evidence in each cluster is used in a first step in force aggregation using templates with more complex relations among several pieces of evidence to form a set of under-determined hypotheses about forces present. In a second step those pieces of evidence not in the core of any cluster are used to complete the under-determined hypotheses where they fit in. This is done by a partial reclustering of non-core intelligence now taking into account the selected templates for each cluster. Finally, the core of each cluster in addition to the newly reclustered evidence of the same cluster are aggregated by templates. By this two-step process the risk of erroneous force aggregation based on some misplaced pieces of evidence from the clustering process is greatly reduced while most of the computation speed may be retained. The new improved algorithm has a computational complexity of $O(n^3 \log^2 n)$ compared to $O(n^2 \log^2 n)$ of standard Dempster-Shafer clustering using Potts spin mean field theory.

## 2 Force aggregation

In this paper our main focus is on using templates in order to improve on the Dempster-Shafer clustering of intelligence reports which in its basic version uses only pairwise conflicts between the reports. We aim to select templates for each cluster based on its core after standard Dempster-Shafer clustering has been performed, and use those templates in a partial reclustering of the intelligence not in the core of any cluster.

However, before focusing on our main problem we must take some preparatory steps. First, we give a short description of standard Dempster-Shafer clustering. For a more extensive description see [5] and [12], [13]. In [12] we describe how to set up the problem, and in [13] how to apply it to intelligence analysis. Furthermore, we use a refined evidence specification [10] in setting up a best possible core [11] of intelligence within each cluster. These preparatory steps are described in Secs. 2.1–2.2 and are included here for completeness.

We then use the core in order to choose a best template for each cluster. When a template has been picked we use it in a partial reclustering to form the basis for a reliable force aggregation. First, we compare the selected template with the non-core and derive for each intelligence report and every cluster the basic belief that it does not belong to the cluster and calculate corresponding interaction, i.e., penalty in the clustering process, between template and non-core intelligence. Secondly, we use the new interactions for a partial reclustering of non-core intelligence. Finally, we aggregate all intelligence in each cluster by the best fitting template. A best fitting template is selected in a slightly different way than during the clustering process. If the reclustering was able to move intelligence into the cluster that fits perfectly with the slots of the initial template not filled up by the core, then we have complete support of that template, and if not, we have some measurable degree of support in this one and in other potential templates. These steps are described in Secs. 2.3–2.7. In Sec. 2.8 we give an example and in Sec. 2.9 we classify forces.

When putting it all together our algorithm for reliable force aggregation performs in seven steps, Figure 1.

| | |
|---|---|
| Step 1: | Perform standard Dempster-Shafer clustering |
| Step 2: | Select the core of each cluster |
| Step 3: | Select a best fitting template for each core |
| Step 4: | Calculate the basic belief against all intelligence in the non-core (and calculate template to intelligence interactions) |
| Step 5: | Use new interactions for a partial reclustering of those intelligence reports not in the core of any cluster |
| Step 6: | Calculate for each cluster a measure of fit between the intelligence of the cluster and all potential templates |
| Step 7: | Aggregate within each cluster using the highest ranking template |

Figure 1: A reliable force aggregation method

## 2.1 Dempster-Shafer clustering

Dempster-Shafer clustering is any method of clustering uncertain data using the conflict in Dempster's rule (or a function thereof) as distance measure.

Several different methods have been developed over the years. Initially methods using iterative optimization was used [14], [1]. While having a good clustering result they had a high computationally complexity. For large scale problems it became clear that we need a method with much lower computational complexity. We then developed faster methods based on clustering using a neural network structure [3], and extended it to simultaneous clustering and determination of number of clusters [4]. The idea to use a neural network structure in this manner were inspired from a solution using neural networks for the traveling salesman problem developed by Hopfield and Tank [15].

Recently a new method [5] using Potts spin [16] mean field theory was developed. This further improves dramatically on the computation time. An extension to this method, setting it up for force aggregation is developed in this paper. A summary of the Potts spin method [5] was included in my paper about managing inconsistent intelligence at FUSION 2000 [13], and a white paper on data management relating to the same was published as [12]. I will in this section provide a short sketch of the method for completeness of this paper.

The clustering method developed may cluster intelligence reports, vehicles and units on all hierarchical levels, Figure 2. Initially, we consider intelligence reports regarding observations of vehicles that come from multiple sources. Here, it is not known apriori if two different intelligence reports refer to the same vehicle.

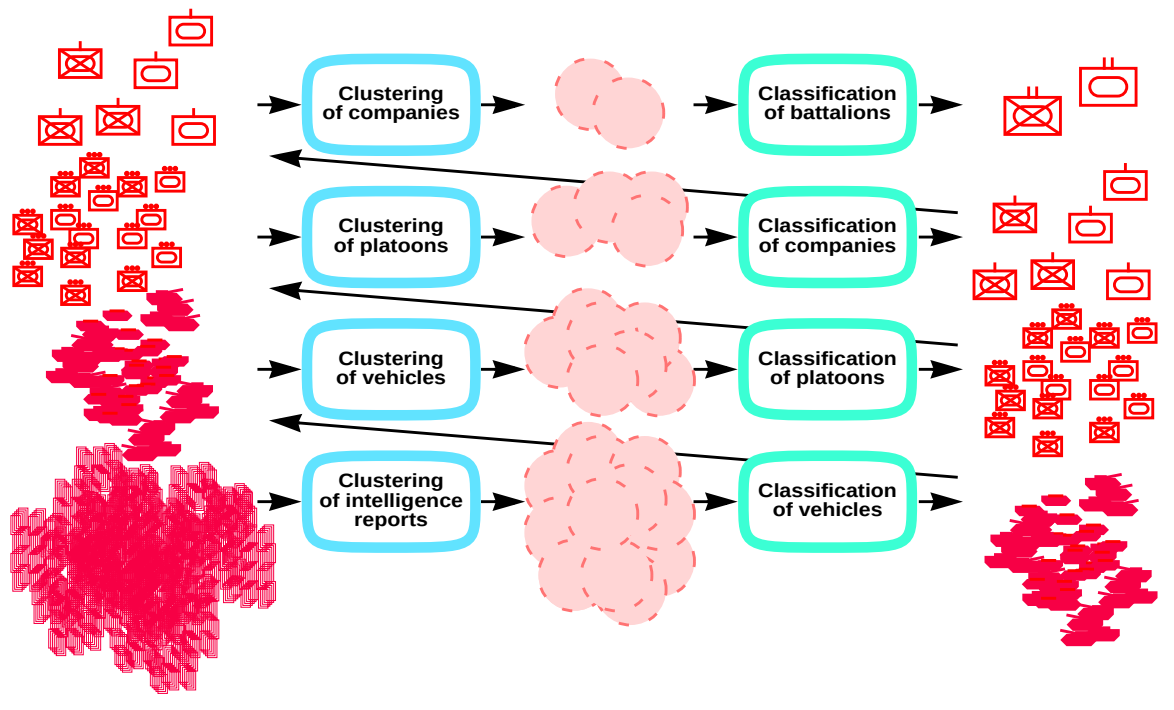

Figure 2: The aggregation process hierarchy

We use the clustering process to separate the intelligence into subsets for each vehicle. We combine Dempster-Shafer theory with the Potts Spin Neural Network model into a powerful solver for large Dempster-Shafer clustering problems.

The problem consists of minimizing an energy function

$$E = \frac{1}{2} \sum_{i,j=1}^{N} \sum_{a=1}^{q} J_{ij} S_{ia} S_{ja} \tag{1}$$

by changing the states of the $S_{ia}$'s, where $S_{ia} \in \{0, 1\}$ is a discrete vector and $S_{ia} = 1$ means that element $i$ is in cluster $a$. This model can serve as a data clustering algorithm if $J_{ij}$ is used as a penalty factor of element $i$ and $j$ being in the same cluster; elements in different clusters get no penalty.

In Dempster-Shafer clustering we use the conflict of Dempster's rule when all elements within a subset are combined as an indication of whether these elements belong together. The higher this conflict is, the less credible that they belong together. To apply the Potts model to Dempster-Shafer clustering we use interactions $J_{ij} = -\log(1 - s_i s_j)\delta_{|A_i \cap A_j|}$. Minimizing the energy function will now minimize the overall conflict.

This minimization process is carried out with simulated annealing. Simulated annealing uses temperature as an important parameter. We start at a high temperature where the $S_{ia}$ change state more or less at random and begin to lower the temperature gradually. Finally when the complete system is frozen, the spins are completely biased by the interactions ($J_{ij}$) so that, hopefully, the minimum of the energy function is reached, giving us a best partition of all intelligence into the clusters with minimal overall conflict.

For computational reasons we will use a mean field model, where spins are deterministic, in order to find the minimum of this energy function. The Potts mean field equations are derived [17] as

$$V_{ia} = \frac{e^{-H_{ia}[V]/T}}{\sum_{b=1}^{K} e^{-H_{ib}[V]/T}} \tag{2}$$

where

$$H_{ia}[V] = \frac{\partial E[V]}{\partial V_{ia}} = \frac{\sum_{j=1}^{N} J_{ij}V_{ja} - \gamma V_{ia} + \alpha \sum_{j=1}^{N} V_{ja}}{G_a} \tag{3}$$

with $V_{ia} \in [0, 1]$. In order to minimize the energy function Eqs. (2) and (3) are used recursively until a stationary equilibrium state has been reached for each temperature. The temperature is lowered step by step by a constant factor until $V_{ia} = 0, 1 \quad \forall i, a$ in the stationary equilibrium state.

The algorithm differs in one respect from the one presented in [5] and [13]. We have included a normalization term [18] within each cluster in Eq. (3)

$$\forall a \; G_a = \frac{K}{N}\sum_{i=1}^{N} V_{ia} . \tag{4}$$

With this normalization we are assured that we can handle problems with different number of elements per cluster. The algorithm described in [5] and [13] required an equal number of elements per cluster.

## 2.2 Finding the core of each cluster

Although by the clustering process every intelligence report was placed in the best cluster for that intelligence report, some reports might belong to one of several different clusters. Such an intelligence report is not useful and should not belong to the core of the cluster.

We must find a measure of the credibility for each intelligence report to belong to the cluster in question. An intelligence report that cannot possibly belong to a cluster has a credibility of zero for that cluster, while an intelligence report which cannot possibly belong to any other cluster and is without any support whatsoever against this cluster has a credibility of one. That is, the degree to which some intelligence report can belong to a certain cluster and no other, corresponds to the importance it wields in that cluster.

The credibility $\alpha_a$ of an intelligence report $C_a^i$ in $\chi_a$ is calculated [11] as

$$\alpha_a = \frac{[\mathrm{Pls}_{\chi_a}(C_a^i \in \chi_a)]^2}{\sum_b \mathrm{Pls}_{\chi_a}(C_a^i \in \chi_b)} . \tag{5}$$

We will use the credibility when determining which intelligence reports among the potential reports for a certain cluster to include in the core of that cluster. Ranking intelligence reports by their credibility eliminates the risk of having some useless intelligence in the core, that could almost have been in the core of some other cluster. We choose intelligence reports with a credibility above some threshold for inclusion in the core. They are judged to be the best representatives of the characteristics of the cluster.

## 2.3 Comparing all cores with all templates

The core of a cluster corresponds to a hypothesis about an element one level higher than the elements of the core, e.g., a number of vehicles imply a specific unit type. In order to find a best template for the core of a cluster we evaluate each available template by comparing the types of the template, e.g., different vehicle types, and the number instances of each type in that template with the same in the entire core.

Let $TY$ be a set of all possible types $\{TY_x\}$; where $TY_x$ is a vehicle or some kind of unit depending on which hierarchical level we are clustering at.

Furthermore, let $T$ be a set of all available templates $\{T_y\}$, and $TK$ a numbered set of $K$ templates selected from $T$, one template for each of $K$ clusters $TK = \{T_a\}, 1 \le a \le K$. Here, some of the $T_a$'s may be equal. This is certainly the case if the set of all available templates $\{T_y\}$ is smaller than the number of clusters; $|\{T_y\}| \le K$.

Each template $T_a$ is represented by any number of slots $S_a^j$ where $S_a^j.pt \in TY$ is a possible type from the set $TY$ and $S_a^j.n$ is the number of that particular type in template $T_a$.

As we allow for nonspecific propositions in the intelligence reports such as “A tank” $\vee$ “An armored personnel carrier” we must calculate the total support for all proposition for both the core and for the template, separately. We calculate this in a fashion similarly to how belief is calculated from the mass functions in Dempster-Shafer theory. For each potential template $T_a$ in cluster $\chi_a$ we calculate the total support $ST_a(\cdot)$ for every subset of $TY$

$$\forall X \in 2^{TY}.\ ST_a(X) = \sum_{i \mid (S_a^i.pt \subseteq X)} S_a^i.n\ . \tag{6}$$

This becomes $\left|2^{TY}\right| - 1$ constraints imposed by each template on the cluster.

Let us make the same analysis for the intelligence. The intelligence reports in cluster $\chi_a$ are divided into two sets, the first set $C_a = \{C_a^i\}$ is the core of intelligence reports, the second set are those that are not in the core of $\chi_a$; $NC_a = \{NC_a^i\}$. Each intelligence report $C_a^i$ in the core of the cluster contains one hypothesis where $C_a^i.pt \in 2^{TY}$ is a subset of all possible types $TY$, and $C_a^i.n$ is the number of that particular subset in intelligence report $C_a^i$. This differs from the representation of the templates that could only accept single types; $S_a^j.pt \in TY$. Also, each intelligence report $NC_a^i$ not in the core of the cluster is structured in the same way as $C_a^i$.

From the set of intelligence reports in the core of the cluster $\{C_a^i\}$ we calculate the total support $SC_a(\cdot)$ in all subsets of the types

$$\forall X \in 2^{TY}.\ SC_a(X) = \sum_{i \mid (C_a^i.pt \subseteq X)} C_a^i.n\ . \tag{7}$$

All subsets supported directly or indirectly (through a sub-subset) by the core of the cluster are constrained by the template in the same cluster such that $SC_a(X) \leq ST_a(X)\ \forall X \in 2^{TY}$. When the constraint holds for all subsets of the types $TY$ we can calculate the admissible number $AC_a(\cdot)$ for each subset of $TY$ that is allowed to be brought into the core of $\chi_a$, through Eqs. (6), (7)

$$\forall X \in 2^{TY}.\ AC_a(X) = ST_a(X) - SC_a(X)\ . \tag{8}$$

$AC_a(X)$ will work as a consistency check for any potential template $T_a$ for the core of $\chi_a$ when $AC_a(X) \geq 0\ \forall X \in 2^{TY}$. If $AC_a(X) < 0$ for any $X \in 2^{TY}$ then the corresponding template is impossible for this cluster as it is overcrowded by the core.

## 2.4 Finding the best template for the core of each cluster

Using $ST_a(\cdot)$ and $SC_a(\cdot)$ we may evaluate all templates by calculating a degree of fit $\mu_{C_a}(T_a)$ between template and core. There are many different possibilities to do this. Here we will take an average between two extremes. The first measure $\mu^1_{C_a}(T_a)$, measures the fit on a type-by-type basis and demands a perfect fit for all types to give a full score. The second measure $\mu^2_{C_a}(T_a)$ ignores all individual types and compares the number of elements of all types in the core with the same in the template. While the first measure seems preferable it can be too extreme when considering missing data for some small number $S_a^j.n$ of slots. Thus, it makes sense to include something like the second measure.

Every template is then evaluated by

$$\mu_{C_a}(T_a) = \begin{cases} \frac{1}{2}[\mu^1_{C_a}(T_a) + \mu^2_{C_a}(T_a)], & \forall X \in 2^{TY}.\ AC_a(X) \geq 0 \\ 0 & \text{otherwise} \end{cases} \tag{9}$$

where the first measure is

$$\mu^1_{C_a}(T_a) = \min\{\mu_{C_a}(S_a^j.pt)\} \tag{10}$$

with

$$\mu_{C_a}(S_a^j.pt) = \frac{SC_a(S_a^j.pt)}{ST_a(S_a^j.pt)}, \tag{11}$$

and the second measure is

$$\mu^2_{C_a}(T_a) = \frac{SC_a(TY)}{ST_a(TY)}. \tag{12}$$

Finally, the preferred template $T_a$ in $\chi_a$ is the template for which we have $\mu_{C_a}(T_a) \geq \mu_{C_a}(T_y)\ \forall\ T_y \neq T_a$.

## 2.5 Finding permitted propositions in the non-core

When clustering the non-core intelligence it is crucial to know what types and how many of each type the template for each cluster asks for. This must be compared with what is currently supported by the core of the cluster, in order to see what possibilities exist for the non-core intelligence.

From the set of intelligence reports $\{NC_a^i\}$ not in the core of the cluster we calculate the total support $SNC_a(\cdot)$ in all subsets of $TY$ in the same manner as in Eq. (7)

$$\forall X \in 2^{TY}.\ SNC_a(X) = \sum_{i \mid (NC_a^i.pt \subseteq X)} NC_a^i.n\ . \tag{13}$$

By comparing the support from intelligence outside of the core $SNC_a(\cdot)$ with that which is admissible to move into the core $AC_a(\cdot)$, we can find how much support in the non-core $NAC_a(\cdot)$ for each subset of $TY$ that is *not* admissible by the template and core to remain in cluster $\chi_a$

$$\forall X \in 2^{TY}.\ NAC_a(X) = max[0, SNC_a(X) - AC_a(X)]. \tag{14}$$

Thus, we have

$$\forall X \in 2^{TY}.\ NAC_a(X) \leq SNC_a(X)\ . \tag{15}$$

## 2.6 Basic belief against non-core intelligence

For each piece of evidence $NC_a^j$ not in the core of cluster $\chi_a$ we define the basic belief $m_{\chi_a}(NC_a^j \notin \chi_a)$ that it does not belong to cluster $\chi_a$. As there are $K$ clusters there are $K$ such basic beliefs for each intelligence report, and a total of $N \times K$ basic beliefs.

As $NAC_a(\cdot)$ is the direct and indirect support for a subset of $TY$ that must not remain in the cluster, out of the $SNC_a(\cdot)$ that are supported, it is the portion $NAC_a(\cdot)$ / $SNC_a(\cdot)$ of the support that must go. Instead of averaging this among all intelligence in the non-core that support the subset directly or indirectly we will focus it entirely on the intelligence supporting the subset directly. This will prioritize specific intelligence before nonspecific, while leaving the final choice of which more specific pieces of intelligence that may remain in the cluster to those $NAC_a(\cdot)$-calculations for the more specific subsets.

If all support for a particular member of $TY$ has an equal chance of being transferred to another cluster that chance must be equal to the same quotient for all such members; $NAC_a(\cdot)$ / $SNC_a(\cdot)$.

**Definition.** Let the basic belief that an intelligence piece $NC_a^j$ does not belong to $\chi_a$ be

$$m_{\chi_a}(NC_a^j \notin \chi_a) = \frac{NAC_a(NC_a^j.pt)}{SNC_a(NC_a^j.pt)}. \qquad (16)$$

One might think that it would be necessary to include a factor $NC_a^j.n$ in the above formula. However, that is not the case. Regardless whether we have many $NC_a^j$'s with $NC_a^j.n = 1$ or one $NC_a^j$ with $NC_a^j.n$ a high number, the basic belief that $NC_a^j$ does not belong to $\chi_a$ should be assigned the same value. In the first case we have same chance of elimination for each $NC_a^j$, and are likely to eliminate a portion of the $NC_a^j$'s. In the second case we have the same chance of eliminating a larger support in one sweep. The expected elimination of support should be the same in both cases.

We may calculate the value of Eq. (16) by Eqs. (6), (7), (8), (13), (14)

$$\begin{aligned} m_{\chi_a}(NC_a^j \notin \chi_a) &= \frac{NAC_a(NC_a^j.pt)}{SNC_a(NC_a^j.pt)} \\ &= \frac{max[0, SNC_a(NC_a^j.pt) - AC_a(NC_a^j.pt)]}{SNC_a(NC_a^j.pt)} \\ &= \frac{max[0, SNC(NC_a^j.pt) - ST(NC_a^j.pt) + SC(NC_a^j.pt)]}{SNC(NC_a^j.pt)} \\ &= max\Bigg[0, \sum_{i|(NC_a^i.pt \subseteq NC_a^j.pt)} NC_a^i.n - \sum_{i|(S_a^i.pt \subseteq NC_a^j.pt)} S_a^i.n \\ &\quad + \sum_{i|(C_a^i.pt \subseteq NC_a^j.pt)} C_a^i.n\Bigg] \Big/ \sum_{i|(NC_a^i.pt \subseteq NC_a^j.pt)} NC_a^i.n \end{aligned}$$

Such basic beliefs are calculated for each intelligence report $NC_a^j$ that is not in the core of any cluster. They will be used in the second clustering.

## 2.7 Second clustering

In order to be able to take into account the impact of templates in the second clustering process we must define additional interactions in each cluster between the template selected for the cluster and all intelligence reports $\{NC_a^i\}$ not in the core of that cluster $\chi_a$. These interactions are different from the previous $J_{ij}$'s that only depended on pairwise conflicts between the intelligence reports themselves. Here we will base the interaction between the template and the intelligence reports on the already derived basic belief $m_{\chi_a}(NC_a^j \notin \chi_a)$ that the intelligence report does not belong to cluster $\chi_a$, Eq. (16).

Let the interaction between template $T_a$ and intelligence reports $NC_a^j$ and $C_a^j$, respectively, be

$\forall 1 \le a \le K, 1 \le j \le N$

$$J_{(N+a)j} = \begin{cases} -\log[1 - m_{\chi_a}(NC_a^j \notin \chi_a)], & \text{for } NC_a^j \\ 0, & \text{for } C_a^j \end{cases}. \qquad (17)$$

That is, an intelligence report not in the core of $\chi_a$ will be penalized by the template whenever $m_{\chi_a}(NC_a^j \notin \chi_a) > 0$, while intelligence reports in the core will never be penalized.

We will bring these interactions into the calculation of the mean field equations, changing Eq. (3) into

$$H_{ia}^s = \frac{\sum_{j=1}^{N} (J_{ij} + \alpha)V_{ja}^s + J_{i(N+a)} + \alpha - \gamma V_{ia}^s}{G_a^s} \quad \forall a. \qquad (18)$$

We now use Eqs. (2) and (18) recursively in the clustering process in the same manner as described in Sect. 2.1. However, Eq. (2) will only be used for intelligence not in the core of any cluster. Thus, when calculating $V_{ia}^s$ (the degree to which intelligence report $i$ belongs to cluster $\chi_a$) during clustering and also when initializing $V_{ia}^0$ before the clustering begins, we will always set $V_{ia}^s$ and $V_{ia}^0 = 1$ if $\exists C_a^i$ (i.e., if intelligence report $i$ is in the core of $\chi_a$), and correspondingly, always set $V_{ia}^s$ and $V_{ia}^0 = 0$ if $\exists C_b^i | b \neq a$ (i.e., if intelligence report $i$ is in the core of some other $\chi_b$, $b \neq a$).

Finally, when $V_{ia}^s$ is larger than some threshold (e.g. $\geq 0.99$) for any $\chi_a$, intelligence report $i$ is moved from the non-core to the core of $\chi_a$ and all $m_{\chi_a}(NC_a^j \notin \chi_a)$ have to be recalculated using Eq. (16). The complete algorithm is shown in Figure 3.

As some of the weights are changed when an intelligence report is moved into the core of a cluster this changes the reclustering problem. Since part of the previous cluster problem is already solved the remaining problem should, hopefully, not be too large. However, while it is not suggested in the algorithm in Figure 3, in a worst case situation we have the option of reinitiating $V_{ia}^0$ (resetting index $s$ = 0) and performing a new clustering. If this is done repeatedly when recalculation of Eq. (16) takes place the computational complexity will increase to $O(n^3 \log^2 n)$.

**INITIALIZE**

$K$ (No. of clusters); $N$ (No. of elements);

$J_{ij} = -\log(1 - s_i s_j)\ \delta_{|A_i \cap A_j|} \quad \forall 1 \le i, j \le N$;

$$J_{i(N+a)} = \begin{cases} -\log[1 - m_{\chi_a}(NC_a^i \notin \chi_a)], & \text{for } NC_a^i \\ 0, & \text{for } C_a^i \end{cases} \quad \text{and}$$

$J_{(N+a)i} = J_{i(N+a)} \quad \forall 1 \le i \le N, 1 \le a \le K$;

$J_{ij} = 0 \quad \forall N+1 \le i, j \le N+K$;

$s = 0$; $t = 0$; $\varepsilon = 0.001$; $\tau = 0.9$; $\alpha$ (for $K \le 7$: $\alpha = 0$, $K = 8$: $\alpha = 10^{-6}$, $K = 9$: $\alpha = 0$, $K = 10$: $\alpha = 3 \cdot 10^{-7}$, $K = 11$: $\alpha = 3 \cdot 10^{-8}$); $\gamma = 0.5$;

$T^0 = T_c$ (a critical temperature) $= \frac{1}{K} \cdot max(-\lambda_{min}, \lambda_{max})$, where $\lambda_{min}$ and $\lambda_{max}$ are the extreme eigenvalues of M, where $M_{ij} = J_{ij} + \alpha - \gamma\delta_{ij}$;

$$V_{ia}^0 = \begin{cases} \frac{1}{K} + \varepsilon \cdot rand[0,1], & \forall b \neg\exists C_b^i \\ 1, & \exists C_a^i \\ 0, & \exists C_b^i \mid b \ne a \end{cases} \quad \forall i, a;$$

**REPEAT**

- REPEAT–2
  - $\forall a\ G_a^s = \frac{K}{N} \sum_{i=1}^{N} V_{ia}$;
  - $\forall i$ Do:
    - $H_{ia}^s = \dfrac{\sum_{j=1}^{N} (J_{ij} + \alpha)V_{ja}^s + J_{i(N+a)} + \alpha - \gamma V_{ia}^s}{G_a^s} \quad \forall a$;
    - $F_i^s = \sum_{a=1}^{K} e^{-H_{ia}^s / T^t}$;
    - $V_{ia}^{s+1} = \begin{cases} \frac{e^{-H_{ia}^s/T^t}}{F_i^s} + \varepsilon \cdot rand[0,1], & \forall b \neg\exists C_b^i \\ 1, & \exists C_a^i \\ 0, & \exists C_b^i \mid b \ne a \end{cases} \quad \forall a$;
  - $s = s + 1$;
  - If $\exists\ 0.99 \le V_{ia}^{s+1} < 1$ then $NC_a^i \to C_a^i$; $V_{ia}^{s+1} = 1$; $V_{ib}^{s+1} = 0\ \forall b \ne a$; $J_{i(N+a)} = 0$; Recalculate $m_{\chi_a}(NC_a^j \notin \chi_a)$ and $J_{j(N+a)}\ \forall j \ne i$ with $C_a^i$.

  UNTIL–2

  $\frac{1}{N} \sum_{i,a} |V_{ia}^s - V_{ia}^{s-1}| \le 0.01$;
- $T^{t+1} = \tau \cdot T^t$;
- $t = t + 1$;

**UNTIL**

$\frac{1}{N} \sum_{i,a} (V_{ia}^s)^2 \ge 0.99$;

**RETURN**

$\{\chi_a \mid \forall S_i \in \chi_a.\ \forall b \ne a\ V_{ia}^s > V_{ib}^s\}$;

Figure 3: Refined clustering

## 2.8 An example

Let us, as an example, perform the calculations of Secs. 2.3–2.6 for one cluster and one template. Assume that we have three different types, $TY$ = {X, Y, Z}, one cluster $\chi_1$ and one template selected for $\chi_1$; $TK = \{T_1\}$. Furthermore, assume the template consists of three slots as follows:

$S_1^1.pt$ = X, $S_1^1.n$ = 4,
$S_1^2.pt$ = Y, $S_1^2.n$ = 2,
$S_1^3.pt$ = Z, $S_1^3.n$ = 2.

Let us also assume that we have nine different intelligence reports and that the standard clustering process has put the five first in the core of $\chi_1$, $C_1 = \{C_1^j\}_{j=1}^5$, and the remaining four in the non-core of the cluster, $NC_1 = \{NC_1^j\}_{j=6}^9$, as follows:

$C_1^1.pt$ ={X}, $C_1^1.n$ = 2,
$C_1^2.pt$ ={X, Z}, $C_1^2.n$ = 1,
$C_1^3.pt$ ={Y}, $C_1^3.n$ = 1,
$C_1^4.pt$ ={Y}, $C_1^4.n$ = 1,
$C_1^5.pt$ ={Z}, $C_1^5.n$ = 1,
$NC_1^6.pt$ = {Y}, $NC_1^6.n$ = 1,
$NC_1^7.pt$ = {X}, $NC_1^7.n$ = 1,
$NC_1^8.pt$ = {X, Y}, $NC_1^8.n$ = 1,
$NC_1^9.pt$ = {Z}, $NC_1^9.n$ = 2.

In Table 1 we calculate the support $\forall X \in 2^{TY}$ from the template $ST_1(\cdot)$, from the core $SC_1(X)$, and that which is admissible to bring into the core $AC_1(X)$, using Eqs. (6), (7) and (8), respectively.

Table 1: Support in different subsets of all types from template and core

| | $ST_1(\cdot)$ | $SC_1(\cdot)$ | $AC_1(\cdot)$ |
|---|---|---|---|
| {X} | 4 | 2 | 2 |
| {Y} | 2 | 2 | 0 |
| {Z} | 2 | 1 | 1 |
| {X, Y} | 6 | 4 | 2 |
| {X, Z} | 6 | 4 | 2 |
| {Y, Z} | 4 | 3 | 1 |
| {X, Y, Z} | 8 | 6 | 2 |

As $AC_a(X) \ge 0$ for all $X \in 2^{TY}$ (Table 1, column three) the template is a possible template for $\chi_1$. Had we had more than one template, we would have ranked them by their degree of fit $\mu_{C_1}(T_y)$ to the core of the cluster. Let us, for the sake of illustration, calculate the degree of fit $\mu_{C_1}(T_1) = \frac{1}{2}[\mu_{C_1}^1(T_1) + \mu_{C_1}^2(T_1)]$ for this template. The first measure is the minimum of all quotients between the third and second column in Table 1; i.e., $1/2$. The second measure is the quotient for {X, Y, Z} of the third and second column; i.e., $6/8$. Thus, $\mu_{C_1}(T_1) = 5/8$, a high fit for $T_1$. As $T_1$ is the only available template in this example it is selected for $\chi_1$.

Continuing with the non-core, we calculate the support $\forall X \in 2^{TY}$ from the non-core $SNC_1(\cdot)$, Eq. (13), and that which is *not* admissible to remain in the cluster $NAC_1(\cdot)$, Eq. (14), in Table 2.

*correction inserted: figure 3*

Table 2: Support in different subsets of all types from non-core

| | $SNC_1(\cdot)$ | $NAC_1(\cdot)$ |
|---|---|---|
| {X} | 1 | 0 |
| {Y} | 1 | 1 |
| {Z} | 2 | 1 |
| {X, Y} | 3 | 1 |
| {X, Z} | 3 | 1 |
| {Y, Z} | 3 | 2 |
| {X, Y, Z} | 5 | 3 |

From the third column, last row, of Table 2 we see that support for three subsets from the non-core must leave the cluster, i.e., the template asks for eight, we have six in the core (Table 1, last row) and five in the non-core, thus three must leave (Table 2, last row).

Using Eq. (16) we calculate the basic belief for all intelligence reports in the non-core, against those remaining in the cluster, Table 3.

Table 3: Basic belief against intelligence belonging to $\chi_1$

| $\{NC_1^j\}$ | $NC_1^j.pt$ | $m_{\chi_1}(NC_1^j \notin \chi_1)$ |
|---|---|---|
| $NC_1^6$ | {Y} | 0 |
| $NC_1^7$ | {X} | 1 |
| $NC_1^8$ | {X, Y} | 1⁄2 |
| $NC_1^9$ | {Z} | 1⁄3 |

Finally, these are used to calculate the interactions for the second clustering using Eq. (17).

## 2.9 Force classification

We will perform force classification individually for each cluster using the best fitting template for that cluster. We will only use one template for each cluster and create one new object for each cluster (corresponding to the object of the template) on the next higher level. Missing and misplaced data in the cluster is tolerated as long as the fit of the best template is high. Any extra misplaced data without corresponding slots in the best template are ignored.

We will here rank the set of all possible templates slightly differently from what we did in Sec. 2.4. Instead of calculating $\mu_{C_a}(T_a)$ for all possible templates we will calculate $\pi_{C_a}(T_a)$ which differs in that we no longer demand that the template is necessarily larger or equal in size to the core. It is sufficient here that it is similar in size to the core, whether larger or smaller. The argument made for the overall size of the template vs. the core is also made for the size of each specific type in the template vs. the same type in the core.

In clustering, one major concern is that of misplaced data. However, in the refined clustering process when using templates in addition to pairwise conflicts between elements our main concern is to allow for missing data in the material. Thus, we assume that any misplaced data will end up in the non-core at this early stage. Now, after the second clustering is finished we must take into account both missing and any misplaced data when performing the final force aggregation. Thus, we might have a core that is slightly smaller than the template (as before in Sec. 2.4) or slightly larger than the template because of misplaced data which ought to belong to another cluster.

Similar to what was done in Sect. 2.4 we evaluate all potential templates $T_a$ for $\chi_a$

$$\pi_{C_a}(T_a) = \frac{1}{2}[\pi^1_{C_a}(T_a) + \pi^2_{C_a}(T_a)] \qquad (19)$$

where

$$\pi^1_{C_a}(T_a) = \min\{\pi_{C_a}(S_a^j.pt)\} \qquad (20)$$

but here we have instead

$$\pi_{C_a}(S_a^j.pt) = \min\left(\frac{SC_a(S_a^j.pt)}{ST_a(S_a^j.pt)}, \frac{ST_a(S_a^j.pt)}{SC_a(S_a^j.pt)}\right) \qquad (21)$$

and

$$\pi^2_{C_a}(T_a) = \min\left(\frac{SC_a(TY)}{ST_a(TY)}, \frac{ST_a(TY)}{SC_a(TY)}\right). \qquad (22)$$

With $\pi^2_{C_a}(T_a)$ as a ranking measure we will slightly favor misplaced data ahead of missing data. This is intuitively correct, as one extra misplaced data piece added to something already perfect will still contain that perfect within, while one missing piece of data away from being perfect is always one piece away. If, for example, $SC_a(TY)$ is five for a particular core in question and $ST_a(TY)$ is four and six for two different templates we will prefer the template with six elements since $5/6 > 4/5$ .

The reliable force aggregation of $\chi = \{\chi_a\}$ is defined as $TK = \{T_a\}$ where $\pi_{C_a}(T_a) > \pi_{C_a}(T_x)\ \forall x \neq a$ for each cluster $\chi_a$, provided that each $\pi_{C_a}(T_a)$ is above some threshold of minimal fit. If it is not, then we have no classification for those $\chi_b$ where $\pi_{C_b}(T_b)$ is below the threshold.

In an recent application oriented paper [19] a simple example with 19 reported vehicles is analyzed, Figure 4.

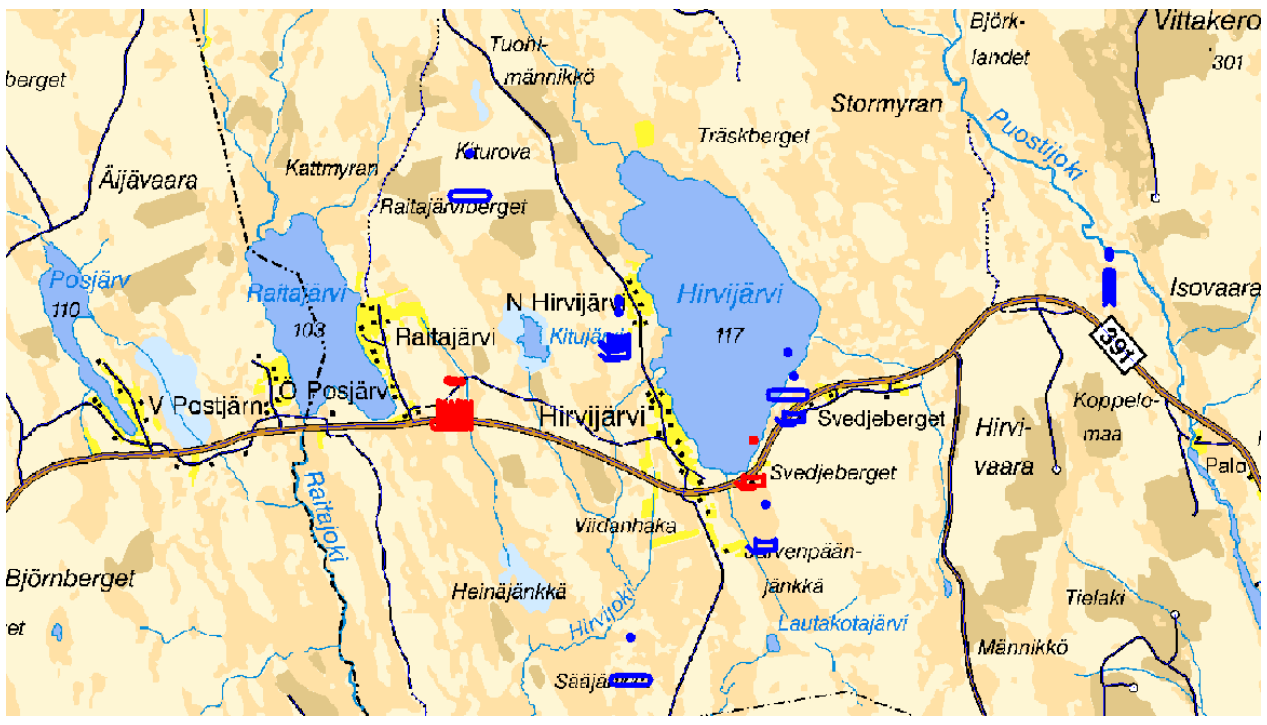

Figure 4: Vehicles

The scenario consists of two opposing forces, with armored personnel carriers (tracked), anti-tank robot vehicles, main battle tanks, and some vehicles only reported as tracked vehicles. Both forces are aggregated. Three templates are used: a “mechanized platoon,” an “anti-tank robot platoon” and a “main battle tank platoon.” Using these templates the 19 vehicles are aggregated into six platoons, Figure 5.

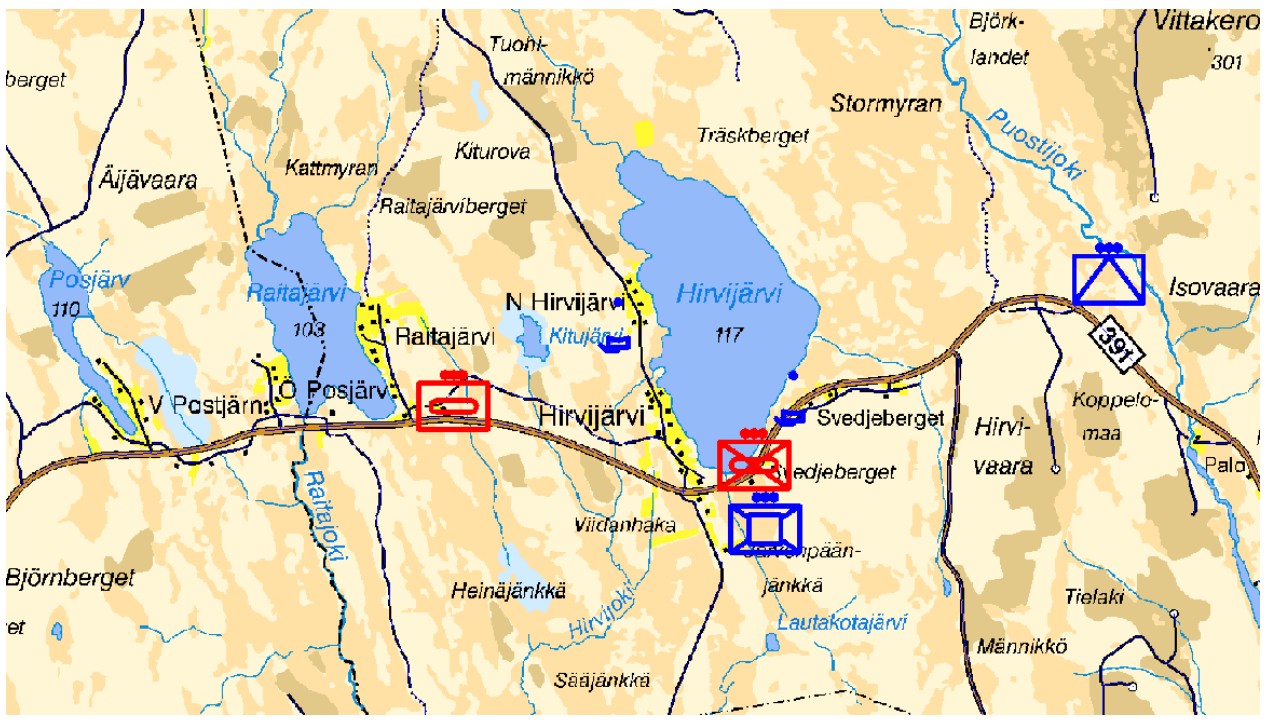

Figure 5: Aggregation result

## 3 Conclusions

The two step clustering process presented in this paper ties together pure clustering and pure classification into an integrated aggregation method. It improves on the accuracy of clustering, with fewer missing or misplaced data pieces. This is accomplished while maintaining most of the excellent computational complexity. This result may form the basis for a more reliable force aggregation.

## References

[1] J. Schubert, Cluster-based specification techniques in Dempster-Shafer theory, in *Symbolic and Quantitative Approaches to Reasoning and Uncertainty, Proc. European Conf.* (ECSQARU'95), C. Froidevaux and J. Kohlas (Eds.), Université de Fribourg, Switzerland, 3–5 Jul 1995, pp. 395–404, Springer-Verlag (LNAI 946), Berlin, 1995.

[2] J. Schubert, Cluster-based Specification Techniques in Dempster-Shafer Theory for an Evidential Intelligence Analysis of Multiple Target Tracks, Ph.D. Thesis, TRITA-NA-9410, ISRN KTH/NA/R--94/10--SE, ISSN 0348-2952, ISBN 91-7170-801-4, Royal Institute of Technology, Stockholm, 1994.

[3] J. Schubert, Fast Dempster-Shafer clustering using a neural network structure, in *Information, Uncertainty and Fusion*, B. Bouchon-Meunier, R.R. Yager and L.A. Zadeh, (Eds.), pp. 419–430, Kluwer Academic Publishers (SECS 516), Boston, MA, 1999.

[4] J. Schubert, Simultaneous Dempster-Shafer clustering and gradual determination of number of clusters using a neural network structure, in *Proc. 1999 Information, Decision and Control Conf.* (IDC'99), Adelaide, Australia, 8–10 Feb 1999, pp. 401–406, IEEE, Piscataway, NJ, 1999.

[5] M. Bengtsson, and J. Schubert, Dempster-Shafer clustering using potts spin mean field theory, *Soft Computing*, to appear.

[6] G. Shafer, *A Mathematical Theory of Evidence*, Princeton University Press, Princeton, 1976.

[7] P. Smets, Practical uses of belief functions, in *Proc. Fifteenth Conf. Uncertainty in Artificial Intelligence* (UAI'99), K.B. Laskey and H. Prade (Eds.), Stockholm, 30 Jul–1 Aug 1999, pp. 612–621, Morgan Kaufmann Publishers, San Francisco, CA, 1999.

[8] P. Smets, Data Fusion in the Transferable Belief Model, in *Proc. Third Int. Conf. Information Fusion* (FUSION 2000), Paris, France, 10–13 Jul 2000, pp. PS/20–33, International Society of Information Fusion, Sunnyvale, CA, 2000.

[9] A. Ayoun, and P. Smets, Data association in multi-target detection using the transferable belief model, *Int. J. Intell. Syst.*, to appear.

[10] J. Schubert, Specifying nonspecific evidence, *Int. J. Intell. Syst.*, Vol 11, No. 8, pp. 525–563, Aug 1996.

[11] J. Schubert, Creating prototypes for fast classification in Dempster-Shafer clustering, in *Qualitative and Quantitative Practical Reasoning, Proc. First Int. Joint Conf.* (ECSQARU-FAPR'97), D.M. Gabbay, R. Kruse, A. Nonnengart and H.J. Ohlbach (Eds.), Bad Honnef, Germany, 9–12 June 1997, pp. 525–535, Springer-Verlag (LNAI 1244), Berlin, 1997.

[12] M. Bengtsson, and J. Schubert, Fusion of incomplete and fragmented data – white paper, FOI-R--0047--SE, Swedish Defence Research Agency, Linköping, 2001.

[13] J. Schubert, Managing inconsistent intelligence, in *Proc. Third Int. Conf. Information Fusion* (FUSION 2000), Paris, France, 10–13 Jul 2000, pp. TuB4/10–16, International Society of Information Fusion, Sunnyvale, CA, 2000.

[14] J. Schubert, On nonspecific evidence, *Int. J. Intell. Syst.*, Vol 8, No. 6, pp. 711–725, Jul 1993.

[15] J.J. Hopfield, and D.W. Tank, "Neural" computation of decisions in optimization problems, *Biol. Cybern.*, Vol 52, pp. 141–152, 1985.

[16] F.Y. Wu, The potts model, *Rev. Modern Physics*, Vol. 54, No. 1, pp. 235–268, 1982.

[17] C. Peterson, and B. Söderberg, A new method for mapping optimization problems onto neural networks, *Int. J. Neural Syst.*, Vol 1, No. 1, pp. 3–22, 1989.

[18] M. Bengtsson, and P. Roivainen, Using the Potts glass for solving the clustering problem, *Int. J. Neural Syst.*, Vol 6, No. 2, pp. 119–132, June 1995.

[19] J. Cantwell, J. Schubert, and J. Walter, Conflict-based Force Aggregation, in *Proc. Sixth Int. Command and Control Research and Technology Symp.*, Annapolis, USA, 19–21 June 2001, to appear.